\documentclass[3p,twocolumn]{elsarticle}

\usepackage{lineno,hyperref}
\modulolinenumbers[5]

\usepackage{amsmath}
\usepackage{amssymb}
\usepackage{color,soul}
\usepackage{tabularx}
\usepackage{MnSymbol}
\usepackage{amsfonts}

\usepackage{url}
\usepackage{times,color,soul}
\usepackage{bm}
\usepackage{multirow}
\usepackage{caption}
\usepackage{subcaption}
\usepackage{afterpage}
\usepackage{arydshln}
\usepackage{url}
\usepackage[shortlabels]{enumitem}
\usepackage{bigstrut}
\usepackage{array}
\usepackage{microtype}
\usepackage{pifont}
\usepackage{textcomp}
\usepackage{siunitx}
\usepackage{flushend}
\usepackage{booktabs,dcolumn}
\usepackage[capitalize]{cleveref}
\usepackage{comment}
\DeclareMathOperator*{\argmax}{\text{argmax}}
\DeclareMathOperator*{\softmax}{\text{softmax}}
\DeclareMathOperator*{\ReLU}{\text{ReLU}}

\journal{Neurocomputing}

\begin{document}

\begin{frontmatter}

\title{Persuasive Dialogue Understanding: the Baselines and Negative Results}

\author[address1]{Hui Chen}
\ead{hui_chen@mymail.sutd.edu.sg}

\author[address1]{Deepanway Ghosal}
\ead{deepanway_ghosal@mymail.sutd.edu.sg}

\author[address1]{Navonil Majumder}
\ead{navonil_majumder@sutd.edu.sg}

\author[address2]{Amir Hussain}
\ead{A.Hussain@napier.ac.uk}

\author[address1]{Soujanya Poria\corref{correspondingauthor}}

\cortext[correspondingauthor]{Corresponding author.}
\ead{sporia@sutd.edu.sg}

\address[address1]{Information Systems Technology and Design, Singapore University of Technology and Design, Singapore}
\address[address2]{School of Computing, Edinburgh Napier University, UK}

\begin{abstract}
Persuasion aims at forming one's opinion and action via a series of persuasive messages containing persuader's strategies. Due to its potential application in persuasive dialogue systems, the task of persuasive strategy recognition has gained much attention lately. Previous methods on user intent recognition in dialogue systems adopt recurrent neural network (RNN) or convolutional neural network (CNN) to model context in conversational history, neglecting the tactic history and intra-speaker relation. In this paper, we demonstrate the limitations of a Transformer-based approach coupled with Conditional Random Field (CRF) for the task of persuasive strategy recognition. In this model, we leverage inter- and intra-speaker contextual semantic features, as well as label dependencies to improve the recognition. Despite extensive hyper-parameter optimizations, this architecture fails to outperform the baseline methods. We observe two negative results. Firstly, CRF cannot capture persuasive label dependencies, possibly as strategies in persuasive dialogues do not follow any strict grammar or rules as the cases in Named Entity Recognition (NER) or part-of-speech (POS) tagging. Secondly, the Transformer encoder trained from scratch is less capable of capturing sequential information in persuasive dialogues than Long Short-Term Memory (LSTM). We attribute this to the reason that the vanilla Transformer encoder does not efficiently consider relative position information of sequence elements.
\end{abstract}


\end{frontmatter}


\section{Introduction}

Persuasive dialogue is an active area of research in the field of dialogue systems and is getting increasing attention from NLP research community recently. In a dyadic persuasive dialogue, there are two interlocutors playing the role of a persuader and a persuadee. The persuader aims to change the persuadee's opinion and reach an intent by using conversational strategies. Although important, there are only a few research studies carried out on persuasive dialogue understanding. Most previous work on persuasiveness mining mainly focuses on the detection and prediction of argumentative features~\cite{ji2018incorporating,chakrabarty2019ampersand}, syntactic features~\cite{tan2016winning} and semantic types of argument components~\cite{hidey2017analyzing} in online persuasive forums.

Persuasive strategies are more complex than ordinary dialogue acts. To identify the persuasive strategies in dialogue, we need a deeper understanding of conversation structures, logical arguments, semantic information of utterances, and even psychological attributes of speakers. Generally, in a persuasive strategy recognition task, each utterance is accompanied by a label containing speakers’ strategy, and then the goal is to identify these strategies by referring to contextual utterances. Hence, this task can be regarded as a sequence labeling task. \cref{tab:snippet} demonstrates a snippet of a persuasive dialogue.  Persuasive strategy recognition is sometimes considered as a subtask of dialogue act recognition~\cite{qindcr,raheja2019dialogue,anikina2019dialogue,chen2018dialogue,ghosal2020utterance}, as they both reflect speakers’ intentions.

Previous methods completely depend on the hidden layers of the network, not accounting for intra-speaker features or self-dependencies that can aid the model with the understanding of logic inertia of individual speakers. These include models that adopt Long Short-Term Memory (LSTM)~\cite{khanpour2016dialogue}, hierarchical LSTM-CNN~\cite{liu2017using}, and hybrid recurrent-CNN~\cite{wang2019persuasion} to extract contextual features and predict labels. However, speakers’ responses will be influenced not only by the semantic history but also by the tactic history.  Naturally, past strategies will influence future strategies. Although these methods have considered contextual correlations in the utterance level, they neglect the accompanied label dependencies in the tactic level. Moreover, intra-speaker dependencies have been neglected. Since the goal of the persuasion dialogue is clear, the persuader usually organizes his/her words strictly and logically during the persuasion process. As we can see in~\cref{tab:snippet}, the persuader carries out two consecutive credibility appeals by two utterances. If we do not look at the previous utterance from the persuader, we can hardly infer which strategy the latter utterance belongs to, as it merely looks like an answer to the persuader's question. Therefore, we conclude that we can improve the model by including intra-speaker features or self-dependencies.

In this paper, we demonstrate the limitations of a Transformer-based approach coupled with Conditional Random Field (CRF) which models contextual features and inter-speaker label dependencies for the task of persuasive strategy recognition. On the benchmark dataset \textsc{PersuasionForGood}~\cite{wang2019persuasion}, our proposed approach presents two negative results. One of them is that the CRF layer does not perform effectively as it does in other tasks such as Named Entity Recognition (NER). We attribute this to the reason that strategies in persuasive dialogues possibly do not follow any strict grammar or rules as the cases in Named Entity Recognition (NER) or part-of-speech (POS) tagging. The other negative result is Transformer-based models do not perform better than LSTM or RNN-based models in this task. We analyze the result and attribute this to the reason that the vanilla Transformer encoder does not efficiently consider relative position information of sequence elements. 

The paper is organized as follows: \cref{related work} discusses the related work on persuasion mining and user intent recognition; \cref{method} elaborates the proposed framework; \cref{experiment} illustrates the experiments; \cref{analysis} shows the results and interprets the analysis, and finally, \cref{conclusion} concludes the paper.

\begin{table*}[ht]
    \centering
    \scriptsize
    \resizebox{\linewidth}{!}{
    \begin{tabular}{lll}
    \hline
    \textbf{Role} & \textbf{Utterance} & \textbf{Annotation} \\
    \hline
     Persuader & Do you ever donate to charity? & task-related-inquiry \\
     Persuadee & Yes, I support a few causes that I personally believe in very much. & positive-to-inquiry \\
     Persuader & Have you ever heard of Save the Children? & source-related-inquiry \\
     Persuadee & Yes, but I don't know a lot about them. & positive-to-inquiry \\
     Persuadee & What is their mission? & ask-org-info \\
     Persuader & Their mission is to promote children's rights, and provide relief and support to children in developing countries. & credibility-appeal \\
     Persuadee & That sounds interesting. & acknowledgement \\
     Persuadee & What countries do they work in? & ask-org-info \\
     Persuader & They work in many countries across the world. & credibility-appeal \\
     Persuader & For example, millions of children in Syria grow up facing the daily threat of violence. & emotion-appeal \\
     Persuader & A donation could help these children greatly. & logical-appeal \\
     Persuadee & It sounds like it. & acknowledgement \\
     Persuadee & Do you donate to this charity? & ask-persuader-donation-intention \\
     Persuader & I do. & self-modeling \\
     Persuader & It is a great charity that does a lot of great work around the world. & logical-appeal \\
     Persuadee & Some charities are run better than others. & other \\
     \hline
    \end{tabular}
    }
    \caption{A snippet of a persuasive dialogue where the annotations include persuasive strategies and non-strategy dialogue acts.}
    \label{tab:snippet}
\end{table*}

\section{Related Work}
\label{related work}
Persuasive communication has been widely explored in various fields such as social psychology, advertising, and political campaigning. To get a better understanding of the persuasiveness of requests on crowdfunding platforms, ~\citet{yang2019let} presented a hierarchical neural network in a semi-supervised fashion to make the persuasiveness quantifiable. \citet{egawa2019annotating} demonstrated five types of elementary units and two types of relations to characterize persuasive arguments and proposed an annotation scheme to capture the semantic roles of arguments in an online persuasive forum~\cite{wei2016post,tan2016winning,hidey2017analyzing}. Furthermore, \citet{hidey2018persuasive} proposed a neural model with words, discourse relations, and semantic frames to predict persuasiveness in social media. Such previous work mainly focuses on evaluating persuasiveness in online forums, neglecting the psychological attributes of different speakers. Hence, in this work, we try to investigate persuasiveness in a conversation setting where persuasion goals, roles of persuader and persuadee as well as interactions between speakers are clearer.

Recent research on user intent recognition has shown promising results. For dialogue act (DA) recognition and classification, \citet{khanpour2016dialogue} presented a deep LSTM structure to classify dialogue acts in open-domain conversations.~\citet{liu2017using} incorporated contextual information for DA classification via a hierarchical deep learning framework. Also, \citet{chen2018dialogue} proposed a CRF-Attentive Structured Network where they captured hierarchical rich utterance representations to help improve DA recognition. For emotion recognition, DialogueRNN~\cite{majumder2019dialoguernn} and DialogueGCN~\cite{ghosal2019dialoguegcn} presented an RNN-based architecture and a GCN-based architecture to grasp hierarchical emotional information and speaker-level dependency. In our task, we try to recognize persuasive strategies utilized in a persuasive dialogue, where not only interactions between speakers make a difference to persuasive strategies but also whether the persuasion succeeds or not has an effect.

\section{Methodology}
\label{method}

\subsection{Problem Definition}

Given two interlocutors persuader and persuadee in a persuasion-driven dialogue $D = (u_1, ..., u_T)$ with $T$ utterances, where utterance $u_t = (w_{t,1}, ..., w_{t,N_t})$ consists of a sequence of $N_t$ words, the goal is to predict the persuasive strategy employed at each utterance. There are 10 and 12 different persuasive strategy categories for the persuader and the persuadee respectively. 
Except for those strategies, there is another category --- `non-strategy dialogue acts' for both the persuader and the persuadee. \cref{tab:labels} presents detailed information about categories of persuasive strategies mentioned in this task. In the remaining sections, we will refer to the persuader and the persuadee as ER and EE respectively.

\begin{table}[ht!]
\centering
\resizebox{\linewidth}{!}{
\begin{tabular}{|l|l|}
\hline
 \multirow[t]{10}{*}{Persuader} & logical appeal \\
(10 categories) & emotional appeal \\
& credibility appeal \\
& foot-in-the-door \\
& self-modeling \\
& personal story \\
& donation information \\
& source-related inquiry \\
& task-related inquiry \\
& personal-related inquiry \\
\hline
\multirow[t]{12}{*}{Persuadee} & ask org info \\
(12 categories) & ask donation procedure \\
& positive reaction \\
& neural reaction \\
& negative reaction \\
& agree donation \\
& disagree donation \\
& provide donation amount \\
& ask persuader donation intention \\
& disagree donation more \\
& task-related inquiry \\
& personal-related inquiry \\
\hline
\multirow[t]{2}{*}{Both Persuader and Persuadee} & non-strategy dialogue acts \\
(1 category common to both) & \\
\hline
\end{tabular}
}
\caption{\label{tab:labels} Categories of persuasive strategies.}
\end{table}

\subsection{Feature Extraction}
\label{ssec:feature-extraction}

We employ the RoBERTa model ~\cite{liu2019roberta} to extract context-independent utterance level feature vectors. RoBERTa is a robustly optimized BERT ~\cite{devlin2018bert} pretraining approach and it uses the same network configuration as BERT which is based upon the widely used Transformer architecture ~\cite{vaswani2017attention}. Several modifications from the BERT pretraining approach is proposed in RoBERTa, which leads to improvement in the end task performance. In particular, there are four key differences in the RoBERTa pretraining approach, which are: i) using dynamic masking instead of static masking, ii) using full sentences without next sentence prediction loss in the next sentence prediction task, iii) using larger mini-batch sizes during training, and iv) using a larger Byte-Pair Encoding (BPE) vocabulary size for tokenization. This modified pretraining procedure results in substantially improved performance in different auxiliary end tasks (GLUE, RACE, and SQuAD).

We fine-tune the RoBERTa Large model for persuasive strategy classification prediction from the transcript of the utterances. RoBERTa Large follows the original BERT Large architecture having 24 layers, 16 self-attention heads in each block, and a hidden dimension of 1024 resulting in a total of 355M parameters. Let an utterance $u_t$ consists of a sequence of BPE tokenized tokens $w_{t,1}, w_{t,2}, ..., w_{t,N_t}$ and its strategy label is $L_t$. In this setting, the fine-tuning of the pretrained RoBERTa model is realized through a sentence classification task. A special token $[CLS]$ is appended at the beginning of the utterance to create the input sequence for the model: $[CLS], w_{t,1}, w_{t,2}, ..., w_{t,N_t}$. This sequence is passed through the model, and the activation from the last layer corresponding to the $[CLS]$ token is then used in a small feedforward network to classify it into its strategy label $L_t$. 

Once, the model is fine-tuned for persuasive strategy classification, we pass the $[CLS]$ appended BPE tokenized utterances to the RoBERTa Large model and extract out activations from the final four layers corresponding to the $[CLS]$ token. These four vectors are then averaged to obtain the context-independent utterance feature vector having a dimension of 1024.

\subsection{Our Model}
 
\begin{figure*}[ht]
\centering
\includegraphics[width=\linewidth]{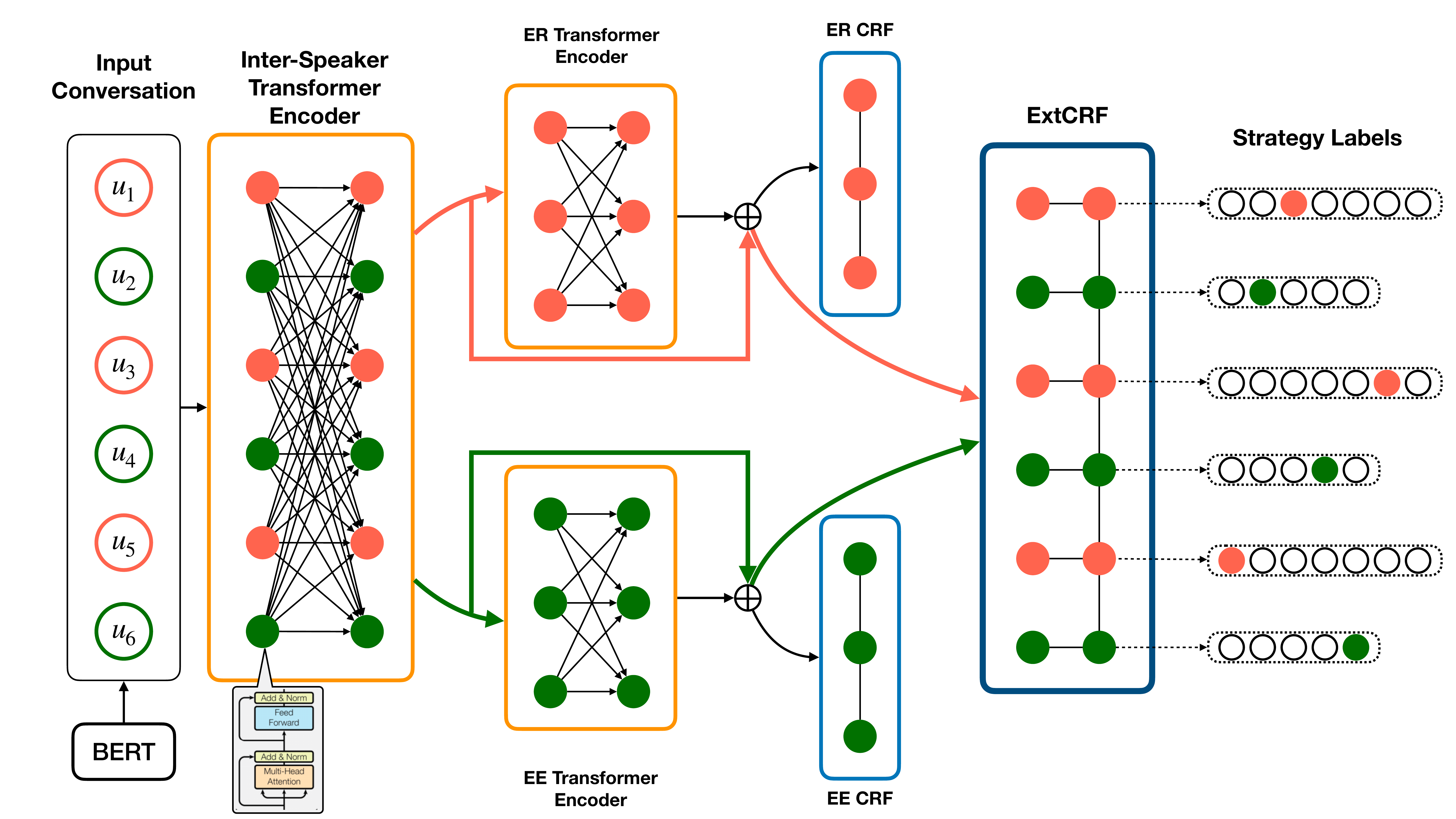}
\caption{\label{fig:architecture} Architecture of our framework. ER and EE represent the persuader and the persuadee respectively, $u$ represents utterance and $\oplus$ represents concatenation operation.}
\end{figure*}

Our model consists of three components: inter-speaker context encoder, speaker-specific context encoder, and strategy classifier. Three Transformers~\cite{vaswani2017attention} first encode both inter- and intra-speaker utterance sequences separately, then these representations are used in a conditional random field (CRF) model~\cite{lafferty2001conditional} to capture label dependencies, and lastly we apply a softmax layer to classify the persuasive strategies. \cref{fig:architecture} shows the architecture of our framework.

\subsubsection{Inter-Speaker Context Encoder}
\label{ssec:sequential-context-encoder}
Persuasive conversations flow along with the responses of a persuader and a persuadee. This sequence contains rich contextual information that can help us better understand the conversation. We feed the whole conversation to a Transformer encoder to capture this inter-speaker contextual information. 

As we illustrated in~\cref{ssec:feature-extraction}, we already obtained the context-independent utterance feature vectors. And the updated utterance representations in each dialogue are composed of these feature vectors. First, these representations $D'= (u'_1, u'_2, ..., u'_T)$ are mapped to queries $Q$, keys $K$ and values $V$ by linear projections with different weights:
\begin{align}
\label{eq:linear-projections}
\begin{split}
    Q_I = W_{q_1}D' \\
    K_I = W_{k_1}D' \\
    V_I = W_{v_1}D' \\
\end{split}
\end{align}
Then, we compute the dot products of the query with all keys to obtain the attention weight, and sum up all the weighted values to produce the context-aware output $Z \in \mathbb{R}^{T\times d_A}$:
\begin{equation}
\label{eq:self-attention}
\begin{aligned}
    Z_I &= Attention(Q_I,K_I,V_I) \\
    &= \softmax(\frac{Q_IK_I^\mathrm{T}}{\sqrt{d_{k_1}}})V_I
\end{aligned}
\end{equation}
where $d_{k_1}$ is the dimension of keys. Further, to enhance the ability of self-attention, we apply multi-head self-attention mechanism here: 
\begin{equation}
    \label{eq:multihead}
    Z_I = [Z_I^{(1)};...;Z_I^{(n)}]W_o \\
\end{equation}
where $Z_I^{(\ast)}$ is computed by~\cref{eq:linear-projections,eq:self-attention},  $[Z_I^{(1)};...;Z_I^{(n)}]$ means the concatenation of all the heads, and $W_o$ is a learnable parameter.

Next, the output $Z$ is fed to a feedforward network which consists of a ReLU activation function and a linear activation function:
\begin{align}
\label{eq:ffn}
\begin{split}
    C_I &= FFN(Z_I) \\
    &= max(0, Z_IW_{f_1}+b_{f_1})W_{f_2}+b_{f_2}
\end{split}
\end{align}
where $W_{\ast}$ and $b_{\ast}$ is the corresponding weight and bias respectively.

\subsubsection{Speaker-Specific Context Encoder}

In a persuasive dialogue, we believe each interlocutor has his/her utterance logic. In this section, we model speaker-specific contextual information. In~\cref{ssec:sequential-context-encoder}, we obtain a new sequential representation and in this section, we separate this sequence into two speaker-specific parts. Here, we define two notations --- 0 represents the persuader and 1 represents the persuadee. Thus, the separated sequences can be written as $C_{I,0} = (u_{0,1}, ..., u_{0,T_{0}})$ and $C_{I,1} = (u_{1,1}, ..., u_{1,T_{1}})$.

Next, like what we have done in~\cref{ssec:sequential-context-encoder}, we feed these two speaker-specific sequences to a Transformer encoder:
\begin{flalign}
\label{eq:speaker-specific-encoder}
    C_{I,0}' &= TrsEncoder(C_{I,0}) \\
    C_{I,1}' &= TrsEncoder(C_{I,1})
\end{flalign}
where the computing way of $TrsEncoder$ is the same as \cref{eq:linear-projections,eq:self-attention,eq:multihead,eq:ffn}.

\subsubsection{Strategy Classification}
We formulate this persuasive strategy classification as a sequence labeling problem. To capture the dependencies among strategy labels, we extend a linear-chain CRF (ExtCRF) to model correlations between labels within neighborhoods in the inter-speaker sequence and do classification.

As we obtain two speaker-specific representations $U_0'$ and $U_1'$ from the speaker-specific encoders, we first concatenate them with the corresponding speaker-specific representations from the inter-speaker Transformer encoder, and next merge these two sequences to one sequence:
\begin{flalign}
    \label{eq:concat0}C_{M,0} &= C_{I,0}' \oplus C_{I,0} \\
    \label{eq:concat1}C_{M,1} &= C_{I,1}' \oplus C_{I,1} \\
    \label{eq:merge-context}C_M &= merge(C_{M,0}, C_{M,1} )
\end{flalign}
where $\oplus$ is the concatenation operation and the $merge(\ast)$ operation merges two speaker-specific sequences $C_{M,0} = (c_{0,1},...,c_{0,T_0})$ and $C_{M,1} = (c_{1,1},...,c_{1,T_1})$ to one sequence $C_{M} = (c_1,...,c_T)$ where $T=T_0+T_1$ and the utterance representations come back to their original positions in the conversation.

\paragraph{ExtCRF classifier}
Next, we feed the merged sequence $C_M$ to our ExtCRF to classify the strategies. Formally, given a sequence of utterances $C_M = (c_1, ..., c_T)$, and the corresponding strategy sequence $Y_M = (y_1,...,y_T)$, the probability of predicting the sequence of strategies can be written as:
\begin{flalign}
\label{eq:crf1}
 P(Y_{M}|C_{M}) &= \frac{1}{\textbf{Z}(C_{M})}\prod_{j=1}^{T}\phi_1(y_{j-1},y_j)\phi_2(y_j, c_j) \\
 \label{eq:crf2}\textbf{Z}(C_{M}) &= \sum_{y' \in \mathcal{Y}}\prod_{j=1}^{T}\phi_1(y_{j-1}',y_j')\phi_2(y_j', c_j)
\end{flalign}
where $\phi_1(\ast)$ and $\phi_2(\ast)$ are feature functions of the state transition potential and the emission potential, respectively. The state transition matrix provides us with the transition scores from label $y_{j-1}$ to label $y_{j}$ and it remains the same for each pair of consecutive time steps. The emission matrix provides us with the scores of label $y_{j}$ at the $j$-th position of the strategy sequence.
\begin{flalign}
\label{eq:crf-ffunc1}
\phi_1(y_{j-1},y_j) &= \exp(W^t_{y_{j-1},y_j}) \\
\label{eq:crf-ffunc2}\phi_2(y_j, c_j) &= \exp(W^e_{y_j}c_j+b^e)
\end{flalign}
where $W^t_{y_{j-1},y_j}$ provides the transition score from label $y_{j-1}$ to label $y_{j}$, $W^e_{y_j}$ maps the context representation $c_j$ to the feature score of label $y_j$, and $b^e$ is the bias of the function.
Different from regular CRF, ExtCRF can deal with multiple label sets of various sizes. In the merged sequence, there are two different types of utterances, one uttered by the persuader and the other uttered by the persuadee. Thus, there exist four state transition cases: ER $\rightarrow$ ER, ER $\rightarrow$ EE, EE $\rightarrow$ ER and EE $\rightarrow$ EE. Accordingly, there are four types of transition matrices where the sizes are $N_r \times N_r$, $N_r \times N_e$, $N_e \times N_r$, and $N_e \times N_e$. $N_r$ and $N_e$ are the total number of labels for the persuader and the persuadee respectively. In our implementation, we integrated these four types of transition matrices into one 4D matrix which contains tag types, and each tag type records a transition matrix.

\paragraph{Normal CRF: ER and EE CRF layers}Except for ExtCRF, here we also adopt a CRF layer to classify the strategies in speaker-specific sequences. Note that in our proposed model, we only take the results of ExtCRF to be the strategy predictions. For these two CRF layers in speaker-specific sequence, we merely add its cross-entropy to the objective function during training. Here the given sequences of utterances are $C_{M,0} = (c_{0,1},...,c_{0,T_0})$ for the persuader and $C_{M,1} = (c_{1,1},...,c_{1,T_1})$ for the persuadee, and the corresponding sequences of predicted labels are $Y_{M,0} = (y_{0,1},...,y_{0,T_0})$ and $Y_{M,1} = (y_{1,1},...,y_{1,T_1})$. Referring to~\cref{eq:crf1,eq:crf2,eq:crf-ffunc1,eq:crf-ffunc2}, we can obtain the probability of predicting the sequence of strategies. There is only one state transition matrix within each CRF layer, and the transition matrices are of size $N_r \times N_r$ for ER CRF and $N_e \times N_e$ for EE CRF, where $N_r$ and $N_e$ are the number of classes in persuader and persuadee labels. 

\paragraph{Persuasion result classification}Further, there is another auxiliary classifier in our completed framework. This classifier aims to predict whether the persuasion succeeds or not. Here we first adopt self-attention to process the dialogue sequence and then apply a two-layer perceptron with a final softmax layer to predict the result:
\begin{flalign}
    l_t &= \ReLU(W_lD_t+b_l) \\
    \mathcal{P}_t &= \softmax(W_{smax}l_t+b_{smax}) \\
    \hat{y_t} &= \argmax_i(\mathcal{P}_t[i]) 
\end{flalign}
where $\hat{y_t}$ is the predicted label for dialogue $D_t$. The cross-entropy of this classifier will be added to the objective function during training.

\subsubsection{Model Training}
We use the sum of cross-entropy from ExtCRF($\mathcal{L}_{m}$), ER CRF($\mathcal{L}_{r}$), EE CRF($\mathcal{L}_{e}$) and persuasion result classifier($\mathcal{L}_{succ}$) along with L2-regularization as the measure of loss($\mathcal{L}$), and our goal is to minimize the objective function during training:
\begin{flalign}
\label{eq:loss-function}
    \mathcal{L} &= \mathcal{L}_{m} + \mathcal{L}_{r} +  \mathcal{L}_{e} +  \mathcal{L}_{succ} + \lambda\left\|\theta\right\|_2 \\
   \label{eq:loss1}\mathcal{L}_{m,r,e} &= -\frac{1}{\sum_{s=1}^N c(s)}\sum_{i=1}^N\sum_{j=1}^{c(i)}\log(P_{i,j}^{m,r,e}[y_{i,j}^{m,r,e}]) \\
   \mathcal{L}_{succ} &= -\frac{1}{\sum_{s=1}^N c(s)}\sum_{i=1}^N\log(P_{i}^{succ}[y_{i}^{succ}]) 
\end{flalign}
where~\cref{eq:loss1} illustrates the computing way of $\mathcal{L}_m$, $\mathcal{L}_{r}$ and $\mathcal{L}_{e}$, N is the number of samples/dialogues, $c(i)$ is the number of utterances in sample $i$, $P_{i,j}^{(\ast)}$ is the probability distribution of predicted labels for utterance $j$ of dialogue $i$, $y_{i,j}^{(\ast)}$ is the expected class label for utterance $j$ of dialogue $i$, $\lambda$ is the L2-regularizer weight, and $\theta$ is the set of all trainable parameters within neural networks. 

Additionally, at the time of testing in CRF layers, we adopt Viterbi algorithm~\cite{viterbi1967error} to obtain the optimal predicted sequence:
\begin{equation}
    \label{eq:crf-test}
    Y^\ast = \argmax_Y(Y|C, \theta)
\end{equation}
where $Y$ is the sequence of predicted labels, $C$ is the sequence of the given sequence of utterances, and $\theta$ is the set of all trainable parameters.

\section{Experimental Setting}
\label{experiment}
\subsection{Dataset}
The dataset used in our experiment is \textsc{PersuasionForGood}~\cite{wang2019persuasion}. There are two types of participants in the dataset. One participant aims to persuade the other participant to donate his/her earning to a charity using different persuasive strategies. It consists of 1017 dialogues, where 300 dialogues are annotated with persuasive strategies. Specifically, there are average 10.43 turns per dialogue and on average 19.36 words per utterance. Also, this dataset provides actual donation made by the persuadee after the session ended. We assess the success of a persuasive dialogue based on whether the persuadee agrees to donate to the charity. In this paper, we use these annotated dialogues to conduct our experiments and partition them into train and test sets with roughly 80/20 ratio. As the dataset is highly imbalanced, here we choose macro F1 to be the evaluation metric. We conduct five-fold cross-validation and take the average scores as the results.

\subsection{Label Dependency}
\label{ssec:label-dependency}
\begin{figure*}[t]
\centering
\includegraphics[width=\linewidth]{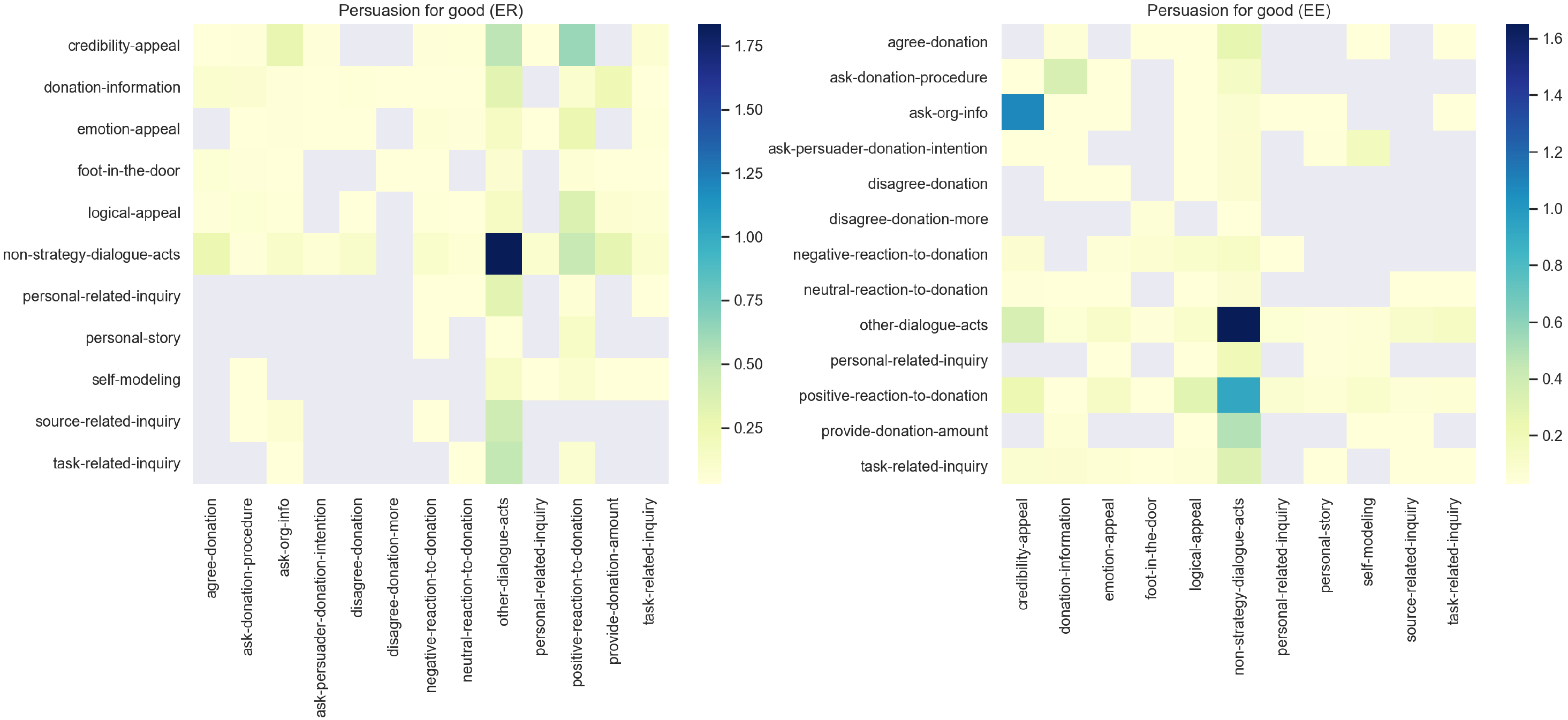}
\caption{The heatmap of inter-speaker label transition statistics in the \textsc{PersuasionForGood} dataset. The left one presents label transitions from persuader strategies to persuadee strategies (ER-to-EE), and the right one is vice versa (EE-to-ER). The color bar represents average number of transitions per dialogue in the dataset.}
\label{Fig.dataset1}
\end{figure*}

\begin{figure*}[t]
\centering
\begin{minipage}{.5\textwidth}
  \centering
  \includegraphics[width=1\linewidth]{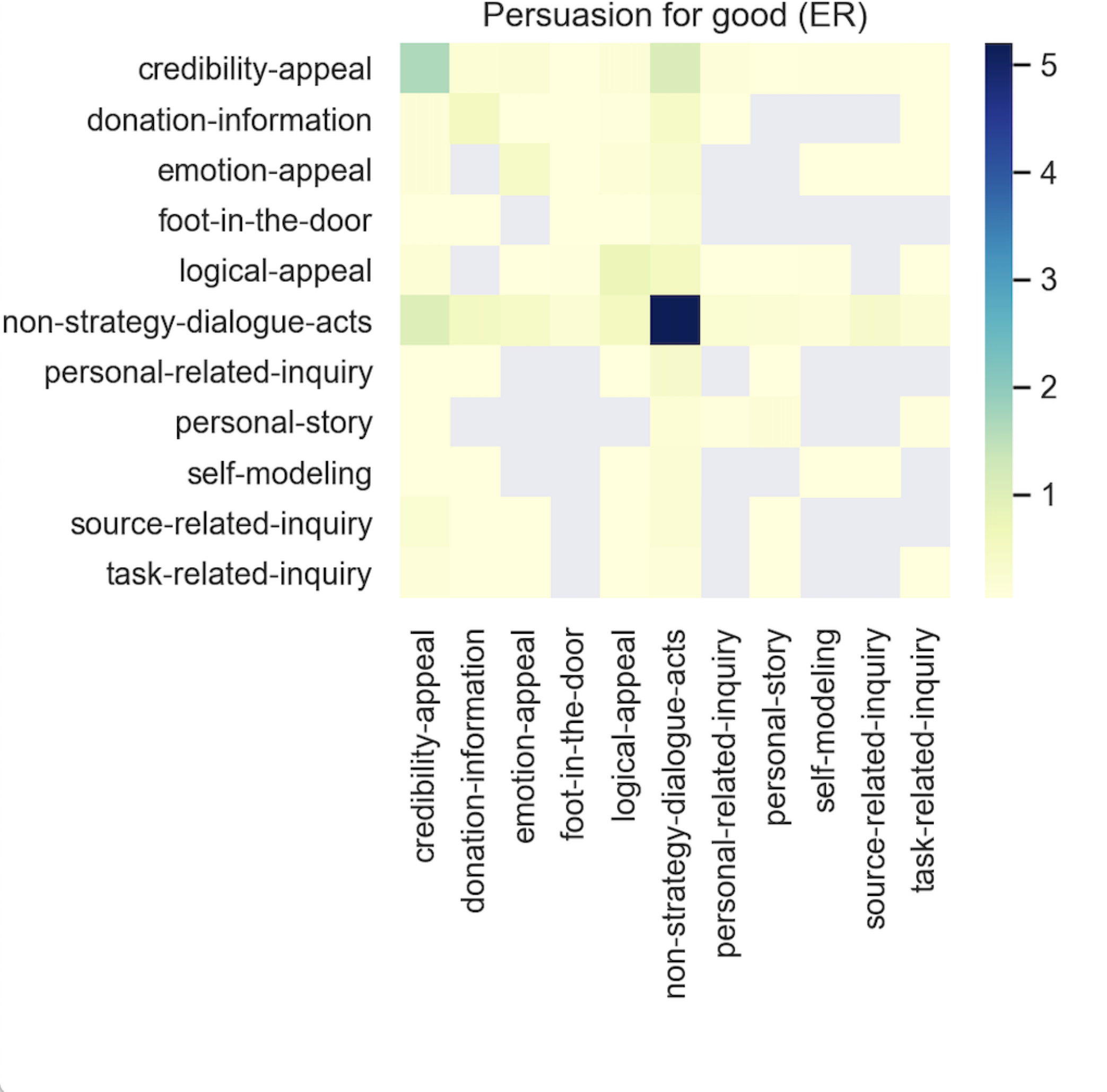}

  \label{Fig.er}
\end{minipage}%
\begin{minipage}{.5\textwidth}
  \centering
  \includegraphics[width=1\linewidth]{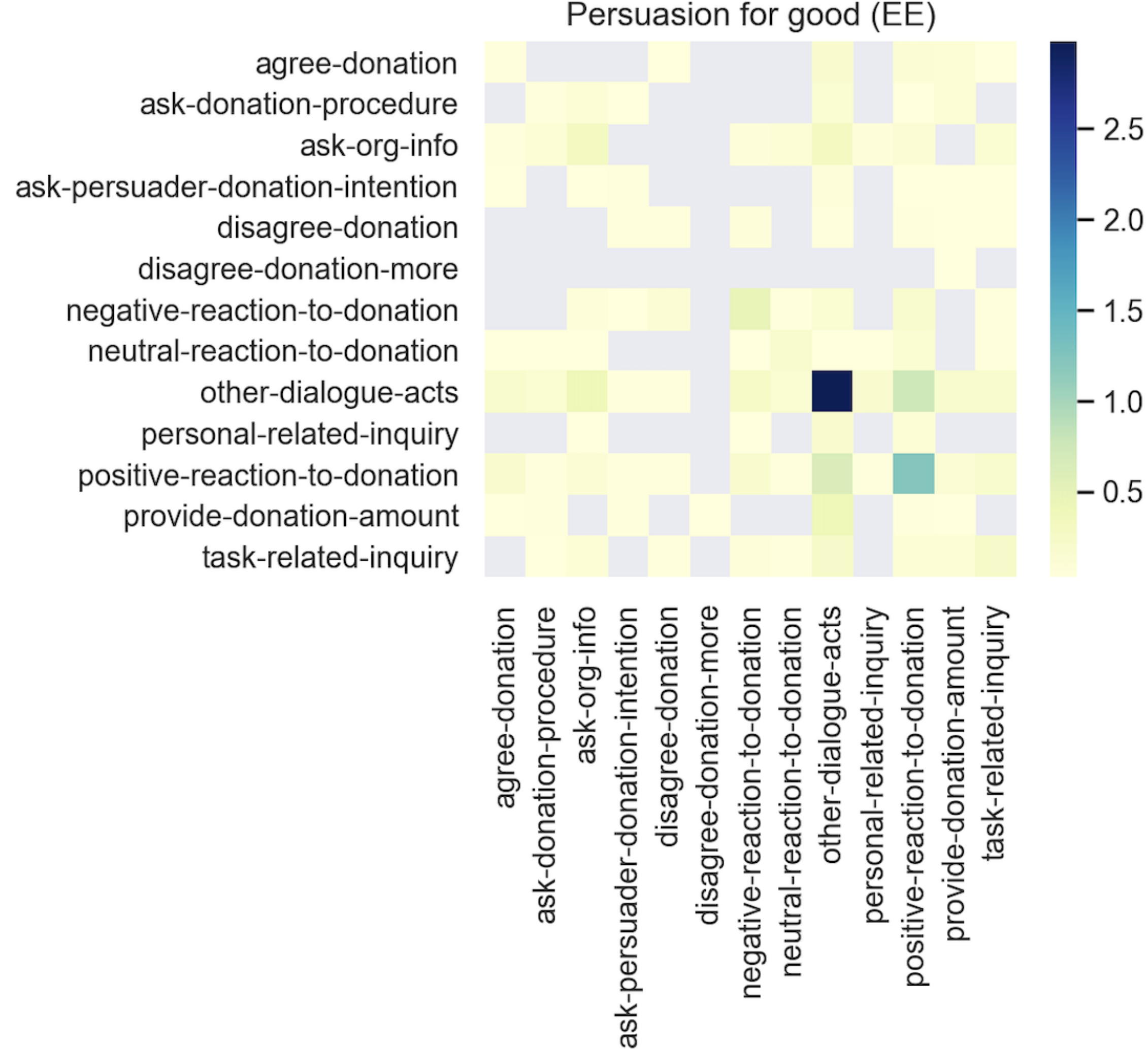}

  \label{Fig.ee}
\end{minipage}
\caption{The heatmap of intra-speaker label transition statistics in the \textsc{PersuasionForGood} dataset. ER and EE present label transitions in persuader's (ER-to-ER) and persuadee's (EE-to-EE) dialogue history, respectively. The color bar represents average number of transitions per dialogue in the dataset.}
\label{Fig.dataset2}
\end{figure*}

To check whether there lies any label dependency in the sequences of the dataset, in~\cref{Fig.dataset1,Fig.dataset2}, we plot frequency of the label pairs $(x, y)$ where $x$ and $y$ are the labels of two consecutive utterances. \cref{Fig.dataset1} presents inter-speaker label transitions and \cref{Fig.dataset2} illustrates intra-speaker label transitions. For both intra- and inter-speaker label transition plots, we can observe that the label pattern with the highest frequency is the combination of two non-strategy dialogue acts. Also, there are other label patterns with high frequency, like (ask-org-info, credibility-appeal) in EE-to-ER label transitions, (credibility-appeal, credibility-appeal) in ER-to-ER label transitions, and (positive-reaction-to-donation, positive-reaction-to-donation) in EE-to-EE label transitions. As we can see in the plots, although there are a couple of label patterns with high frequency in this dataset, most of other patterns are with low frequency. Hence, the dependency characteristic of labels is not obvious in this dataset.

\subsection{Baselines}

To obtain a comprehensive evaluation, we compare the proposed model with the following baseline methods:

\paragraph{RoBERTa LogReg~\cite{liu2019roberta}}
As RoBERTa has achieved good performance in many NLP tasks, we take RoBERTa as a baseline where we feed the utterance-level feature vectors obtained by RoBERTa to a fully connected multi-layer perceptron layer to perform the classification. The input features are non-contextual as there is no information flow from contextual utterances. For simplicity, we call this model RoBERTa LogReg (Logistic Regression).

\paragraph{RoBERTa cLSTM~\cite{poria2017context}, bcLSTM~\cite{poria2017context}}
Besides RoBERTa LogReg, we also present results on two widely used sequence-based models contextual LSTM (cLSTM) and bidirectional contextual LSTM (bcLSTM). Similar to RoBERTa LogReg, we first use RoBERTa~\cite{liu2019roberta} to train the embeddings of the whole conversation, then feed the conversation to a cLSTM/bcLSTM, and lastly we adopt a two-layer perceptron with a final softmax layer to predict the strategies. Contextual LSTM creates context-aware utterance representations by capturing the contextual features from the surrounding utterances using an LSTM. Bidirectional contextual LSTM is similar to cLSTM but without bidirectionality in the LSTM module. 

\paragraph{RoBERTa DialogueRNN~\cite{majumder2019dialoguernn}}
DialogueRNN is a recurrent neural network-based model that aims to model inter-speaker relations and can be applied to multiparty datasets. It uses two gated recurrent units (GRUs) to track individual speaker states and global context in the conversation.

\paragraph{cLSTMs, cLSTMs-CRF, cLSTMs-ExtCRF~\cite{hochreiter1997long,lafferty2001conditional}}
The models cLSTMs, cLSTMs-CRF and cLSTMs-ExtCRF have employed the same architecture of our proposed model, which leverages both inter- and intra-speaker contextual features. The only key difference between cLSTMs with our model is the former uses LSTM instead of transformer. Compared to cLSTMs, cLSTMs-CRF adds a CRF layer to the speaker-specific contextual encoders. And based on cLSTMs-CRF, cLSTMs-ExtCRF adds another CRF layer to the inter-speaker encoder. The output of the last layer of these models is fed to a two-layer perceptron with a final softmax layer to predict the strategies.

\paragraph{DialogueRNN-cLSTMs-CRF, DialogueRNN-cLSTMs-ExtCRF~\cite{majumder2019dialoguernn,hochreiter1997long,lafferty2001conditional}}
These two models keep the same structure of cLSTMs-CRF and cLSTMs-ExtCRF. The difference is that we adopt DialogueRNN to encode the inter-speaker contextual features.

\paragraph{Transformer-based Models~\cite{vaswani2017attention}}
Attention mechanism has been widely used in various NLP tasks in recent years. In our baselines, we adopt Transformers to encode both the conversation flow and the speaker-specific utterances. For the Transformers baseline, we directly feed the outputs of Transformers to a two-layer perceptron with a final softmax layer and obtain the prediction results. For Transformers with CRF layers, we add a CRF layer to the Transformer encoders. Transformers-cLSTMs-ExtCRF is a baseline that we change the speaker-specific contextual encoder in our proposed method from Transformers to contextual LSTMs.

\section{Results and Analysis}
\label{analysis}

We compare our model with baseline methods for persuasive strategy classification in~\cref{tab:results-text}. As the dataset is highly imbalanced, we select macro F1 to be the evaluation metric. Due to the paucity of annotated data, we conduct five-fold cross-validation and use the averaged scores as the final results for one training run. Due to the variances in training convergence we encountered, we performed each experiment 5 times. We take the average of all 5 runs and report the average score in~\cref{tab:results-text}. For Transformer-based models, we set the learning rate to be 0.00001 and L2 regularization weight to be 0.00001, and for RNN-based models, the learning rate is 0.0001. The batch size is 16 and each model is trained for 65 epochs. Moreover, we utilize the validation set to tune the hyper-parameters. More details about hyper-parameters can be found in~\cref{tab:hyper-params}.

\begin{table*}[t]
\centering
\begin{tabular}{lcc}
\toprule 
Setting & Transformer-based models & RNN-based models  \\ 
\midrule
 batch size & 16 & 16 \\
 optimizer & Adam & Adam \\
 learning rate & 1e-5 & 1e-4 \\
 L2 regularization weight & 1e-5 & 1e-5 \\
 dropout rate & 0.1 & 0.1 \\
 training epochs & 65 & 65 \\
 \# attention heads & 2 & - \\
 \# layers & 2 & 1 \\
 hidden size & 1024 & 1024 \\
 token embedding dimension & 1024 & 1024 \\
\bottomrule
\end{tabular}
\caption{\label{tab:hyper-params} Hyper-parameter details in our experiments.}
\end{table*}

\subsection{Comparisons and Negative Results}
In all our baselines, word embeddings are trained by RoBERTa. Firstly, we observe that contextual models perform better than the non-contextual Logistic Regression model. \cref{tab:results-text} shows the model with a contextual LSTM encoder obtains a 0.8\% F1-score increase in the persuader strategy prediction and a 1.4\% F1-score increase in the persuadee strategy prediction, compared with RoBERTa LogReg. Similarly, when we use a Bidirectional LSTM or some RNN layers to encode the conversation sequence, the performances are both better than that of RoBERTa LogReg. Secondly, the results show that the intra-speaker contextual feature improves the Macro F1 scores. All the baseline models containing both inter- and intra-speaker contextual features perform better than those without intra-speaker contextual features. In particular, for the models using LSTM to encode the intra-speaker contextual features, the F1 scores in the persuadee strategy prediction are all above 52\%. Compared with RoBERTa cLSTM, cLSTMs model obtain obvious F1 increases in both persuader and persuadee strategy predictions. This demonstrates that explicitly capturing intra-speaker dependencies contributes to the persuasive strategy recognition.

\noindent However, in these experiments, we also observe two negative results:
\begin{enumerate}
    \item Adding CRF layers does not improve the results. We think this phenomenon is possibly because strategies in persuasive dialogues do not follow any strict grammar or rules. Also, \cref{ssec:label-dependency} shows the frequency of label dependencies is low in this dataset, which makes CRF hard to capture dependency features.
    \item Transformer-based models do not perform better than LSTM-based models. Although Transformers have an ability to capture long-term dependencies, it is not as effective as LSTM in this task. As we know, the self-attention mechanism utilized in vanilla Transformers is unaware of positions. They use position embeddings generated by sinusoids of varying frequency~\cite{vaswani2017attention} to record the position information. However, this kind of position encoding approach does not efficiently model the relative positions or distances between sequence elements~\cite{shaw2018self}. Moreover, we think the paucity of annotated data makes the position information hard to be learned in Transformers.
\end{enumerate}

\begin{table*}
\centering
\begin{tabular}{lcccc}
\toprule \multirow{2}{*}{\textbf{Models}} & \multicolumn{2}{c}{\textbf{Persuader}} & \multicolumn{2}{c}{\textbf{Persuadee}} \\
 & W-Avg F1 & Macro F1 & W-Avg F1 & Macro F1 \\ 
\midrule
 RoBERTa LogReg & 73.9 & 63.3 & 63.6 & 50.5 \\
 RoBERTa cLSTM & 75.2 & 64.1 & 66.2 & 51.9 \\
 RoBERTa bcLSTM & 75.1 & 64.0 & 66.4 & 51.7 \\
 RoBERTa DialogueRNN & 75.1 & 64.3 & 65.8 & 51.8 \\
\midrule
cLSTMs & 75.5 & \textbf{65.5} & 66.3 & \textbf{52.5} \\
cLSTMs-CRF & 75.6 & 65.2 & 66.5 & 52.3 \\
cLSTMs-ExtCRF & 75.6 & 65.3 & 66.3 & 52.3 \\
DialogueRNN-cLSTMs-CRF & 75.1 & 64.6 & 65.9 & 52.1 \\
DialogueRNN-cLSTMs-ExtCRF & 75.1 & 64.5 & 66.2 & 52.2 \\
\midrule
Transformers & 74.8 & 64.6 & 65.5 & 51.4 \\
Transformers-CRF & 74.8 & 64.8 & 65.2 & 51.5 \\
Transformers-ExtCRF (this work) & 75.0 & 65.2 & 65.4 & 51.6 \\
Transformers-cLSTMs-ExtCRF & 75.1 & 65.3 & 66.0 & 51.4 \\
\bottomrule
\end{tabular}
\caption{\label{tab:results-text} Comparison with the baseline methods on \textsc{PersuasionForGood} dataset. W-Avg F1 and Macro F1 represent weighted average F1 score and Macro F1 score. The unit of the scores is \%.}
\end{table*}

\subsection{Performance In the BI-Scheme Label Setting}
\begin{figure*}[t]
\centering
\includegraphics[width=0.95\linewidth]{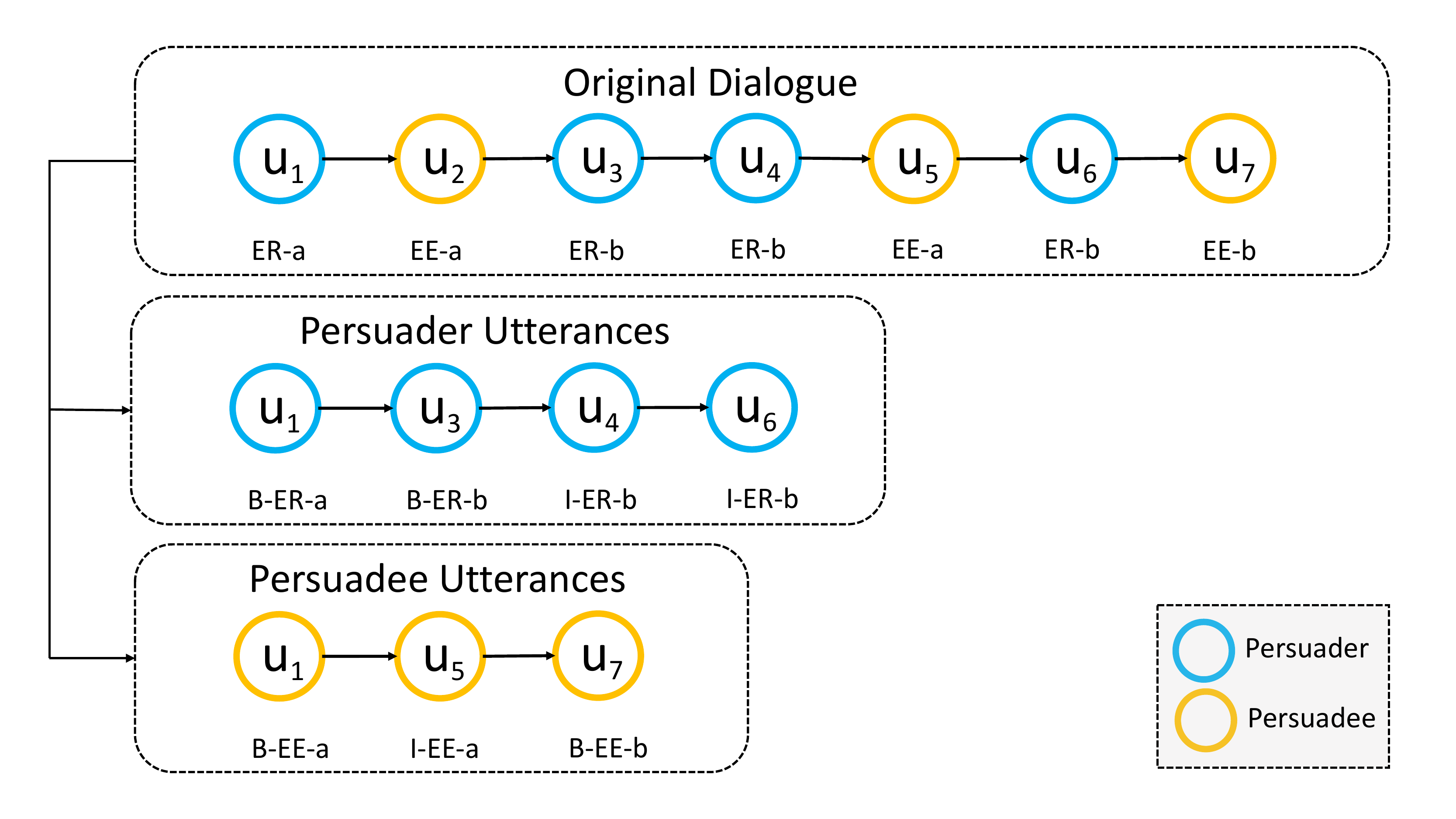}
\caption{A diagram of the BI-scheme Labels. Blue and orange circles represent utterances of the persuader and the persuadee, respectively. ER-a and ER-b stand for two types of labels in persuader strategies and the prefixes 'B' and 'I' are the symbols for Begin scheme and Input scheme.}
\label{Fig.BI-scheme}
\end{figure*}

To further investigate the performance of our model, we conduct a couple of experiments in the BI-scheme label setting, where we separate each type of label into two schemes: Begin scheme and Input Scheme. \cref{Fig.BI-scheme} shows how we annotate the BI-scheme labels. In the BI-scheme label setting, if the current label is the same as the previous label in the original dialogue, the current label is an Input-scheme label, otherwise, it is a Begin-scheme label. \cref{tab:bi-results} presents the performances of different models in the BI-scheme label setting. As we see in~\cref{tab:bi-results}, the proposed model surpasses all the baselines in the persuader strategy prediction, where the Macro F1 score achieves 65.2\%. For the persuadee strategy prediction, cLSTMs-ExtCRF performs best, with the Macro F1 score achieving 52.7\%. Although we have manually identify the boundaries between label spans, there is no obvious and stable improvement in models with CRF.

\begin{table*}[t]
\centering
\begin{tabular}{lcccc}
\toprule \multirow{2}{*}{\textbf{Models}} & \multicolumn{2}{c}{\textbf{Persuader}} & \multicolumn{2}{c}{\textbf{Persuadee}} \\
 & W-Avg F1 & Macro F1 & W-Avg F1 & Macro F1 \\
\midrule
 RoBERTa cLSTM & 74.1 & 63.4 & 66.0 & 51.5 \\
 RoBERTa bcLSTM & 74.8 & 64.4 & 66.0 & 51.9 \\
 RoBERTa DialogueRNN & 74.7 & 64.2 & 65.6 & 51.9 \\
 \midrule
cLSTMs & 75.1 & 64.5 & 66.4 & 52.2 \\
cLSTMs-CRF & 75.1 & 64.6 & 66.5 & 52.3 \\
cLSTMs-ExtCRF & 75.2 & 64.7 & 66.6 & \textbf{52.7} \\
\midrule
Transformers & 74.7 & 64.9 & 65.7 & 52.3 \\
Transformers-CRF & 75.0 & 64.6 & 65.7 & 52.0 \\
Transformers-ExtCRF (this work) & 75.1 & \textbf{65.2} & 65.9 & 51.8 \\
\bottomrule
\end{tabular}
\caption{\label{tab:bi-results} Comparison with the baseline methods on \textsc{PersuasionForGood} dataset in the BI-scheme label setting. W-Avg F1 and Macro F1 represent weighted average F1 score and Macro F1 score. The unit of the scores is \%.}
\end{table*}

\subsection{Case Studies}
In this section, we analyze the predictions of our model and the predictions of the Transformers. In~\cref{tab:case study}, we list some cases to compare our method with the Transformers. When encountering utterances that contain very little semantic information, e.g., non-strategy dialogue acts, our model maintains a good performance while the Transformers model does not. Moreover, we found our model performs better in the recognition of credibility appeal strategy combinations. There are several such strategy combinations in the dataset and they usually appear after the `ask org info' strategy from the persuadee. Generally in such combinations, the first one mainly replied to the persuadee and gave the information he/she asked, and the second one is what the persuader intended to express. In this case, semantic information alone is not enough. And as our proposed model considers the strategy transition, the predictions improve a lot. Further, there are some strategies like `disagree-donation' and `negative-reaction' that have something in common and are easy to be confused. In this case, the contextual information plays an important role in distinguishing such labels. 

However, we also observed some weaknesses in our model. In some cases, the persuadee was not willing to donate at first, but after persuasion, he/she agreed. Neither our model nor the Transformers baseline has achieved satisfactory results. In most of these cases, there is a long distance between the `disagree-donation' attitude and the `agree-donation' attitude in a dialogue. Our model does not perform well in capturing long-distance strategy dependencies.

\begin{table*}[ht]
    \centering
    \scriptsize
    \resizebox{\linewidth}{!}{
    \begin{tabular}{lccc}
    \hline
     \textbf{Utterance} & \textbf{Gold label}&\textbf{Pred. of Transformers-ExtCRF} & \textbf{ Pred. of Transformers} \\
    \hline
     ER:By directly asking for aid. & neutral-to-inquiry (Non) & Non & logical-appeal \\
     EE:Thank you for your time. & thank(Non) & Non & disagree-donation \\
     EE:What kind of children's charities do you know about? & task-related-inquiry & task-related-inquiry & ask-org-info \\
     ER:Some of the causes they support include Emergencies (38\%), Health & & & \\
       and Nutrition (36\%), and Education to more than 136 thousand children & & & \\
       all over the world. & credibility-appeal & credibility-appeal & logical-appeal \\
     ER:I am supposed to ask you if you care about people being killed & & & \\
     in Syria and things like that, I don't want to cause you any & & & \\
     emotional discomfort by talking about suffering people. & emotional-appeal & emotional-appeal & logical-appeal \\
     EE:I would like to donate \$0 but its not because I don't believe in the cause. & disagree-donation & disagree-donation & negative-reaction \\
    
    \hline
    \end{tabular}
    }
    \caption{Samples in case studies. `Non' represents non-strategy dialogue acts. ER and EE represent the persuader and the persuadee respectively.}
    \label{tab:case study}
\end{table*}

\begin{figure*}[t]
\centering
\begin{minipage}{.5\textwidth}
  \centering
  \includegraphics[width=1\linewidth]{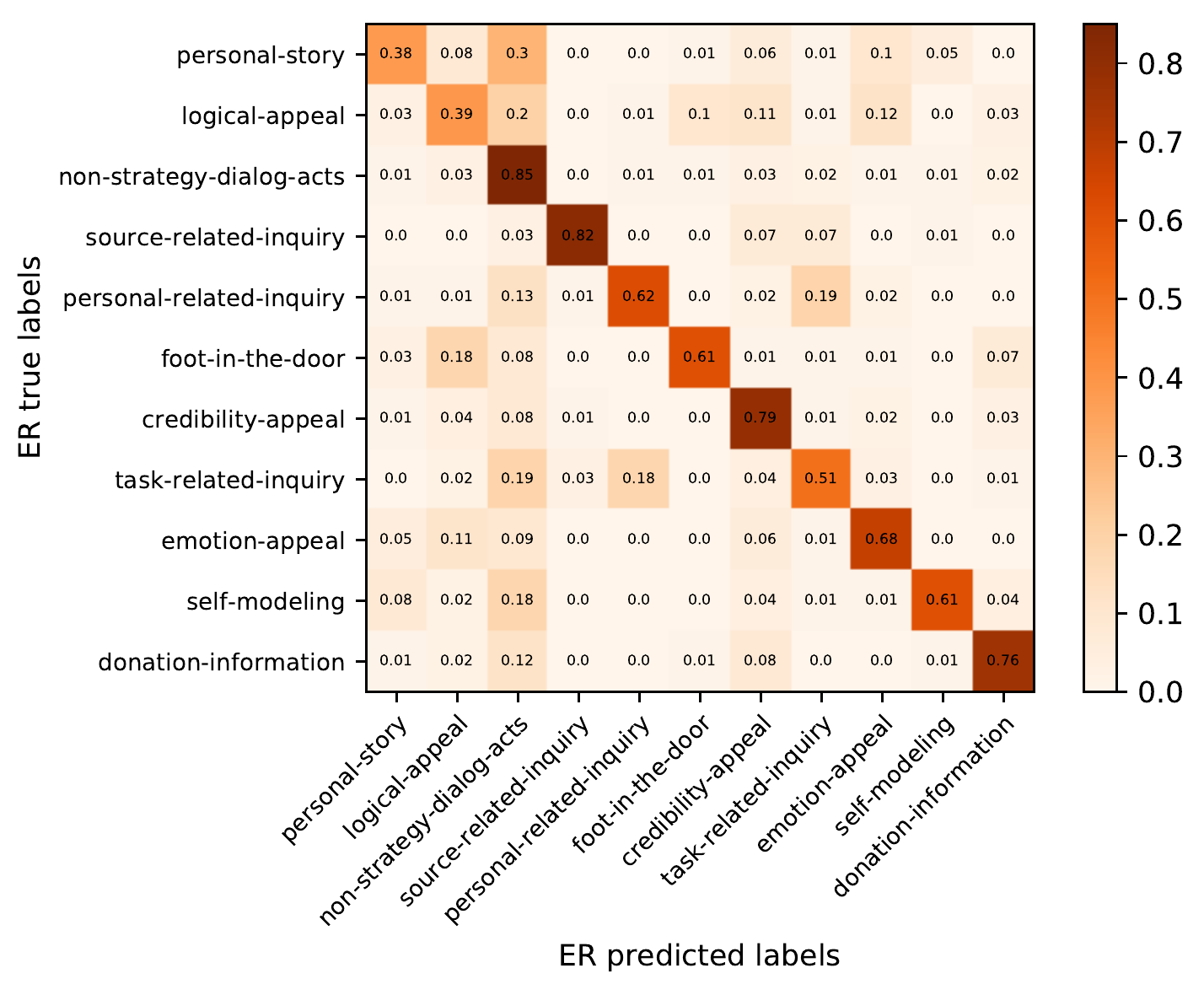}

  \label{Fig.sub.1}
\end{minipage}%
\begin{minipage}{.5\textwidth}
  \centering
  \includegraphics[width=1\linewidth]{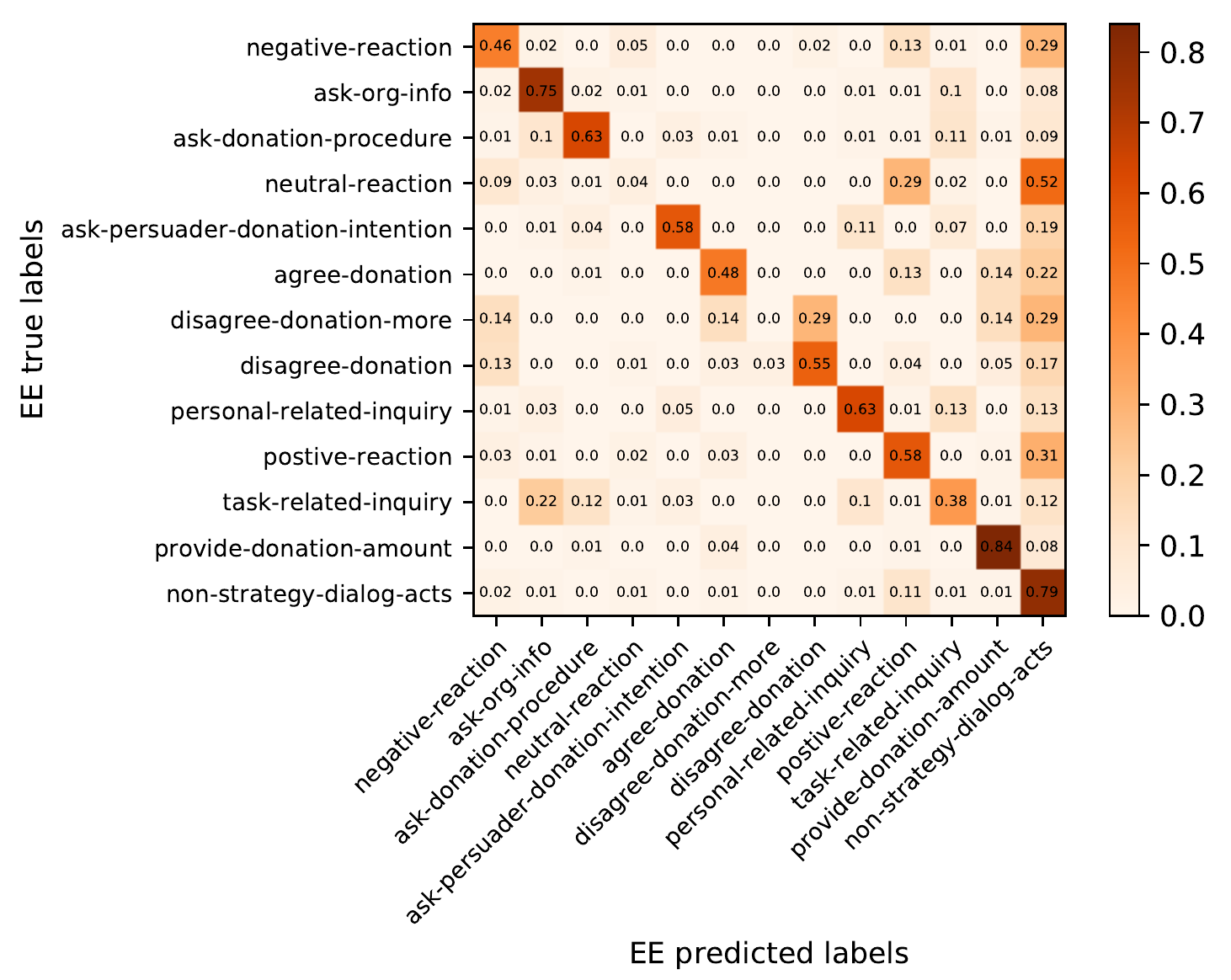}

  \label{Fig.sub.2}
\end{minipage}
\caption{Confusion matrix of our model for (a) persuader strategy classification, and (b) persuadee strategy classification.}
\label{Fig.main}
\end{figure*}

\subsection{Error Analysis}
As shown in~\cref{Fig.main}, we visualize the performance of our proposed model in two confusion matrices. In~\cref{Fig.sub.1}, we observed that `personal story' tends to be misclassified into `non-strategy dialog acts'. This is because utterances telling personal stories usually present an inconspicuous strategy tendency. Further, we found that several samples of `logical appeal' are misclassified as `emotional appeal' and `credibility appeal'. One of the reasons is that one utterance may have multiple appeals. For instance, \textit{`Save the Children is able to give away nearly everything they gather.'} This utterance can be classified into logical appeal since it tells the persuadee if he/she donates, the organization will probably help many young children. Also, it can be classified into credibility appeal since the organization tries to earn the persuadee's trust via this utterance.

Moreover, we observed there are more samples misclassified as `non-strategy-dialog-acts' in persuadee strategy classification as shown in~\cref{Fig.sub.2}. For instance, the majority of samples of `neutral-reaction' are misclassified as `non-strategy-dialog-acts'. Similarly, one reason is that neutral reaction usually presents an inconspicuous strategy tendency. Further, we found samples of `disagree-donation-more' are easily misclassified as `disagree-donation'. We surmise this is due to the subtle difference between these two labels. Our model leaves some room for improvement to distinguish very similar labels.

\section{Conclusion}
\label{conclusion}
In this paper, we first introduce a Transformer-based neural network coupled with extended CRF, that captures both inter-speaker and intra-speaker contextual features and label dependencies to recognize persuasive strategies in dialogues. And then, through a couple of experiments on the benchmark dataset, we compare the proposed approach with several baselines and obtain two negative results that help us get a deeper understanding of persuasive dialogues. Future work will focus on generating diverse persuasive responses to enhance the ability of non-collaborate dialogue agents.

\bibliographystyle{elsarticle-num-names}
\bibliography{main}

\end{document}